\documentclass[conference]{IEEEtran}
\IEEEoverridecommandlockouts

\usepackage{cite}
\usepackage{amsmath,amssymb,amsfonts}
\usepackage{graphicx}
\usepackage{caption}
\usepackage{subcaption}
\usepackage{xcolor}
\usepackage{booktabs}
\usepackage{amsthm}
\usepackage{framed}
\usepackage{gradient-text}
\usepackage{xspace}
\usepackage[hidelinks]{hyperref}

\definecolor{remarkbox}{rgb}{0.9, 0.95, 1.0}
\newenvironment{remark}{%
    \MakeFramed{\advance\hsize-\width\FrameRestore}%
}{\endMakeFramed}

\newcommand{\caustream}{\gradientRGB{CauSTream}{92,154,216}{255,140,0}\xspace}

\newtheorem{assumption}{Assumption}

\newtheorem{corollary}{Corollary}

\newcommand{\modelName}[1]{\texttt{CauSTream}}

\begin{document}

\title{%
  \raisebox{-0.5ex}{\includegraphics[height=30pt]{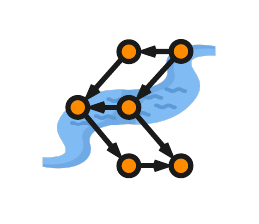}}\hspace{-0.2em}%
  \caustream: Causal Spatio-Temporal Representation Learning for Streamflow Forecasting
}

\author{
\IEEEauthorblockN{Shu Wan\IEEEauthorrefmark{1}, Reepal Shah\IEEEauthorrefmark{2}, John Sabo\IEEEauthorrefmark{2}, Huan Liu\IEEEauthorrefmark{1}, K. Sel\c{c}uk Candan\IEEEauthorrefmark{1}}
\IEEEauthorblockA{
    \IEEEauthorrefmark{1}\textit{School of Computing and Augmented Intelligence, Arizona State University} \\
    \{swan, huanliu, candan\}@asu.edu \\
    \IEEEauthorrefmark{2}\textit{The ByWater Institute, Tulane University}\\
    \{rshah3, jsabo1\}@tulane.edu
    }
}

\maketitle

\begin{abstract}
Streamflow forecasting is crucial for water resource management and risk mitigation. While deep learning models have achieved strong predictive performance, they often overlook underlying physical processes, limiting interpretability and generalization. Recent causal learning approaches address these issues by integrating domain knowledge, yet they typically rely on fixed causal graphs that fail to adapt to data. 
We propose \caustream, a unified framework for causal spatiotemporal streamflow forecasting. CauSTream jointly learns (i) a runoff causal graph among meteorological forcings and (ii) a routing graph capturing dynamic dependencies across stations. We further establish identifiability conditions for these causal structures under a nonparametric setting.
We evaluate CauSTream on three major U.S. river basins across three forecasting horizons. The model consistently outperforms prior state-of-the-art methods, with performance gaps widening at longer forecast windows, indicating stronger generalization to unseen conditions. Beyond forecasting, CauSTream also learns causal graphs that capture relationships among hydrological factors and stations. The inferred structures align closely with established domain knowledge, offering interpretable insights into watershed dynamics. CauSTream offers a principled foundation for causal spatiotemporal modeling, with the potential to extend to a wide range of scientific and environmental applications.
\end{abstract}

\begin{IEEEkeywords}
Causal representation learning, Causal discovery, Spatio-temporal learning, Hydrologic modeling, Streamflow forecasting
\end{IEEEkeywords}

\section{Introduction}
Accurate streamflow forecasting is critical for water resource management, supporting applications from flood control to hydropower scheduling and ecosystem preservation. Process-based models such as VIC-CMF~\cite{liang1994simple, yamazaki2011physically} simulate streamflow through a two-stage mechanism: meteorological \emph{forcings} (e.g., precipitation, temperature) generate surface and subsurface \emph{runoff}, which is then routed through river networks to produce downstream \emph{streamflow}. While grounded in physical principles, these models require extensive basin-specific calibration and are infeasible to deploy in regions with limited data. In the past decade, deep learning models such as ConvLSTM~\cite{shi2015convolutional} and Spatio-Temporal Graph Convolutional Networks (STGCN)~\cite{yu2017spatio} have shown strong predictive performance \cite{kratzert2018rainfall}. However, these models often behave as ``black boxes,'' offering limited interpretability and reduced robustness under distribution shifts.

A recent promising direction is the causal machine learning model, where a directed acyclic graph (DAG) encodes causal knowledge to guide model learning. For instance, Causal Streamflow Forecasting (CSF)~\cite{wan2024csf} utilizes a river flow network as the causal graph, improving prediction accuracy. However, such approaches depend on a \emph{pre-defined} causal graph, instead of \emph{learning} it from data. Recent advances in nonlinear causal discovery show that such causal structures can be identified from observational data under mild assumptions \cite{khemakhem2020variational}, opening the door to end-to-end models that jointly perform causal discovery and prediction.

We introduce CauSTream, a framework for \underline{Cau}sal \underline{S}patio\underline{T}emporal st\underline{ream} forecasting. CauSTream is a general causal learning framework that integrates causal discovery with multi-step streamflow prediction. Inspired by process-based hydrological models, it learns two causal graphs: (i) an instantaneous DAG ($\mathcal{G}_F$) that captures dependencies among meteorological forcings, and (ii) a spatiotemporal routing DAG ($\mathcal{G}_Q$) that governs runoff routing. The learned causal graphs provide both physical interpretability and improved generalization in streamflow forecasting.

CauSTream is implemented in two variants designed for efficiency and adaptability:
\begin{enumerate}
    \item \textbf{CauSTream-Shared}, which learns a single global runoff function suitable for relatively homogeneous catchments;
    \item \textbf{CauSTream-Local}, which uses a hypernetwork \cite{ha2017hypernetworks} to capture station-specific heterogeneity, analogous to calibration in hydrological models.
\end{enumerate}

We evaluate CauSTream on three large U.S. river basins (Brazos, Colorado, and Upper Mississippi). The empirical results demonstrate substantial improvements over conventional forecasting methods, affirming the capability of our causal framework to effectively model complex hydrological processes. Thus, our work contributes significantly to advancing streamflow prediction methodologies and providing robust, interpretable insights into hydrological dynamics across diverse geographic contexts.

Our contributions are as follows:
\begin{itemize}
    \item We propose CauSTream, a novel end-to-end framework that, to our knowledge, is the first to unify causal discovery and multi-step-ahead forecasting for streamflow prediction;
    \item We demonstrate state-of-the-art forecasting performance on three large-scale, hydrologically diverse U.S. river basins (Brazos, Colorado, and Upper Mississippi), consistently outperforming strong baselines, particularly on long-range horizons;
    \item We show that our model learns interpretable causal graphs for both meteorological forcings ($\mathcal{G}_F$) and streamflow routing ($\mathcal{G}_Q$) that align with established hydrological principles and domain knowledge, and
    \item We validate that the model's internal runoff embedding is aligned with the runoff simulated by the hydrological model.
\end{itemize}

\section{Background and Related Work}\label{review}

\begin{figure}[htp]
    \centering
    \includegraphics[width=\linewidth]{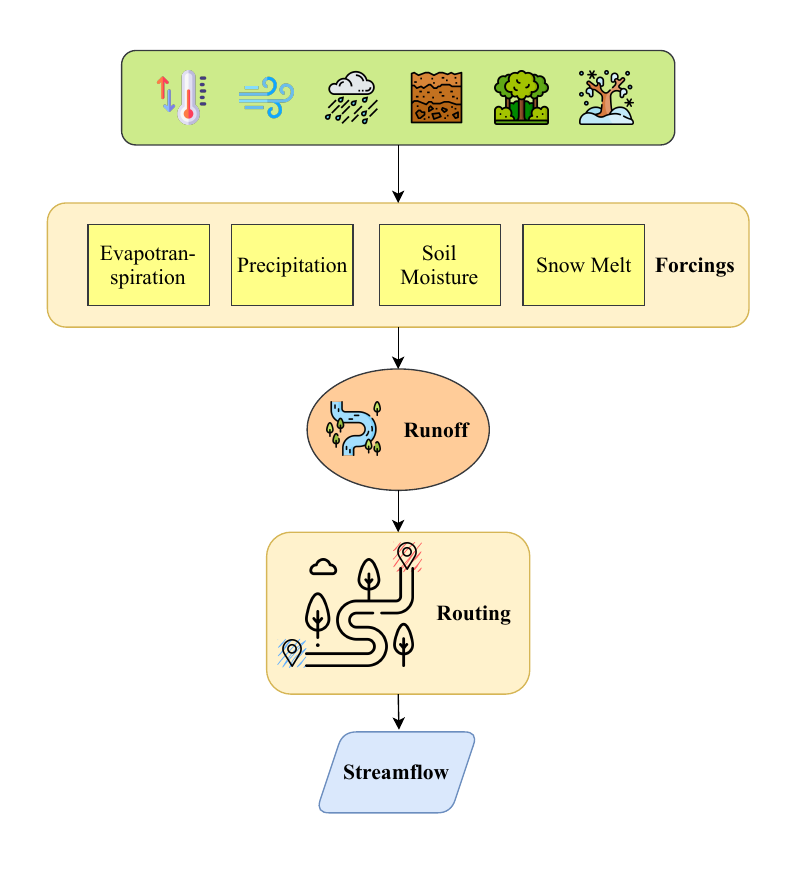}
    \caption{VIC-runoff-CaMa-Flood workflow.}
    \label{fig:vic-cmf}
\end{figure}

\textbf{Process-based hydrologic models}. The Variably Infiltration Capacity (VIC) model \cite{liang1994simple} is a widely used large-scale distributed model that assumes (i) universal governing equations for hydrologic processes and (ii) independent treatment of each grid with locally calibrated parameters. While this structure enables flexibility, it introduces bias by neglecting cross-grid interactions. In coupled modeling frameworks, runoff is the sole output passed from VIC to the Catchment-based Macro-scale Floodplain (CaMa-Flood) model \cite{yamazaki2011physically}, which performs river routing to estimate streamflow (Fig.~\ref{fig:vic-cmf}). This VIC–runoff–CaMa-Flood configuration has become a standard approach and has demonstrated high fidelity in prior studies \cite{ wang2021modeling}. However, despite its accuracy, process-based models demand extensive calibration and detailed geophysical inputs, making them time-consuming and difficult to apply at scale or for real-time forecasting.

\textbf{Causal-guided machine learning models}. Machine learning methods such as Convolutional Neural Networks, Bayesian Neural Networks, Graph Neural Networks, and Conv-LSTM networks have demonstrated strong predictive performance in hydrologic forecasting \cite{shi2015convolutional, kratzert2018rainfall, liu2023applying, deng2022deep}. However, purely data-driven methods often fail to capture inherent causal relationships such as upstream-downstream connectivity and drainage hierarchy, leading to weaker predictive performance under unseen conditions \cite{chang2023artificial}. Physics-informed and hybrid approaches attempt to mitigate these shortcomings by embedding physical principles into machine learning frameworks \cite{shen2018transdisciplinary}, yet challenges remain in disentangling rainfall-runoff generation from routing processes \cite{nearing2021role, shen2018transdisciplinary}.  

To address these limitations, causal-guided methods explicitly model causal dependencies in spatiotemporal data~\cite{christiansen2022toward}. Standard STGCNs typically rely on geographical or distance-based adjacency, often misrepresenting the true causal structure in river networks. Recent causal-guided approaches attempt to incorporate explicit domain knowledge or perform causal discovery to infer accurate causal structures. For example, Wan et al. \cite{wan2024csf} leverage river flow networks derived from DEM as explicit causal priors for streamflow forecasting. Alternatively, when such causal structures are unknown, causal discovery methods become necessary. Sheth et al. \cite{sheth2023streams} use causal discovery algorithms and reinforcement learning to infer causal relationships from observed data. Despite these advancements, integrating explicit causal discovery into hydrologic forecasting remains challenging, largely due to the complexity of spatiotemporal interactions and limited theoretical foundations in causal inference for nonlinear hydrologic systems.

\textbf{Causal discovery and identifiability.} Traditional causal discovery approaches, such as LiNGAM~\cite{shimizu2006linear} often requires linearity and thus inadequate for capturing the nonlinear dynamics systems. Recent advances extend causal discovery to nonlinear settings by leveraging connections to nonlinear Independent Component Analysis (ICA)~\cite{zheng2022identifiability,hyvarinen2019nonlinear,khemakhem2020variational}, where causal structure can be inferred from the support of the Jacobian matrix of learned mixing functions. Fu et al.~\cite{fu2025ncdl} established theoretical guarantees for identifying causal graphs in nonlinear systems under the assumptions of functional faithfulness, sufficient variability, and independent, non-Gaussian noise. Building on these insights, our framework explicitly incorporates hydrological domain knowledge via DEM-derived masks to constrain causal discovery, significantly reducing computational complexity and ensuring physically consistent causal structures.

\begin{figure*}[htp]
    \centering
    \includegraphics[width=.8\linewidth]{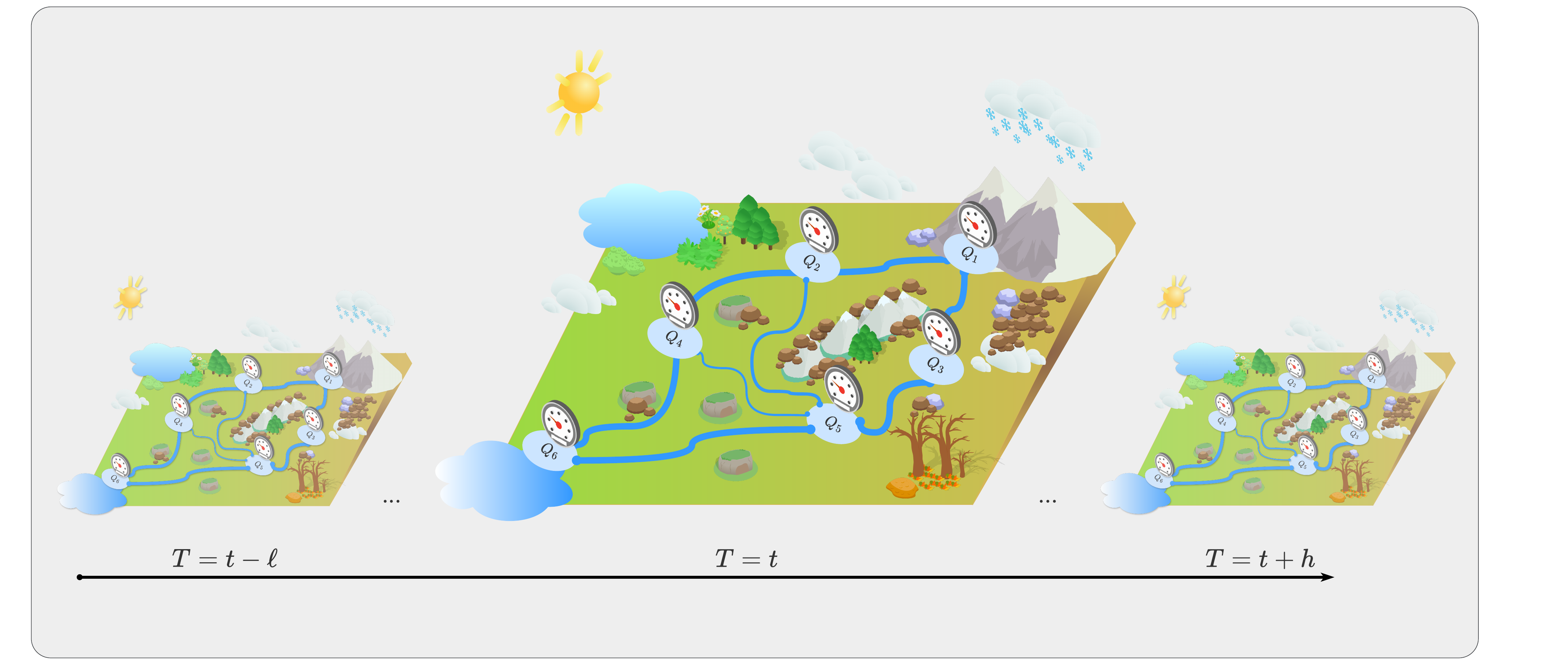}
    \caption{Illustration of the spatiotemporal streamflow forecasting task. Given a history window of forcings and streamflow observations ($T=t-\ell$ to $T=t$), 
the goal is to predict streamflow at future timesteps ($T=t+h$) across multiple stations within the river network.}
    \label{fig:streamflow-diagram}
\end{figure*}
\section{Problem Formulation}\label{problem}

In this section, we formally define the spatiotemporal causal graph forecasting problem. Inspired by streamflow forecasting tasks traditionally approached using hydrologic models like VIC-runoff-CMF, we first introduce general notations for spatiotemporal forecasting (Section~\ref{subsec:stg}). Then, we instantiate these notations to clearly define the streamflow data-generating process (DGP) underlying our proposed method (Section~\ref{subsec:streamflow-dgp}). Finally, we formalize the identifiability conditions and theoretical basis of causal inference for this DGP (Section~\ref{subsec:identifiability}).

\subsection{Spatiotemporal Causal Graph Forecasting}\label{subsec:stg}
Consider $N$ spatial locations indexed by $\mathcal{N}=\{1,\dots, N\}$, and $t\in \{1,\dots,T \}$ discrete time steps. At each location $k$ and time $t$, we observe an input vector of covariates $\mathbf{F}_{t,k}\in\mathbb{R}^{d_f}$ and a scalar target variable $Q_{t,k}\in\mathbb{R}$. We stack inputs and targets across stations to form matrices $\mathbf{F}_t\in\mathbb{R}^{N\times d_f}$ and $\mathbf{Q}_t\in\mathbb{R}^{N}$. Let $\ell \in \{0, 1, \cdots, L\}$ denote the history length used for forecasting.  

A Spatiotemporal Causal Graph Forecasting task is defined with a spatiotemporal directed acyclic graph (STDAG) $\mathcal{G}=\langle\mathcal{V},\mathcal{E}\rangle$, where the node set is 
\[
\mathcal{V} = \{\mathbf{Q}_{t-\ell:t}, \mathbf{F}_{t-\ell:t}\},
\] 
including target variables and covariates across locations with temporal indices backtracked up to $\ell$. The graph encodes both instantaneous and lagged causal dependencies that govern the data-generating process.

Given a historical context window of length $\ell$ and a forecast horizon $H$, the forecasting task learns a function $f$:
\begin{equation}\label{eq:forecast_general}
    \hat{\mathbf{Q}}_{t+H}=f\bigl(\mathbf{F}_{t-\ell:t},\mathbf{Q}_{t-\ell:t},\mathcal{G}\bigr),
\end{equation}
such that predictions $\hat{\mathbf{Q}}_{t+h}$ approximate the true targets $\mathbf{Q}_{t+h}$ for $1 \leq h \leq H$, while respecting the spatiotemporal causal dependencies entailed by $\mathcal{G}$.

In the context of streamflow forecasting, the covariates $\mathbf{F}$ correspond to meteorological forcings and the targets $\mathbf{Q}$ correspond to streamflow. 
We model this forecasting task as a Spatiotemporal Causal Graph Forecasting problem, where the causal graphs are designed to reflect hydrologic processes. 

\subsection{Hydrology-guided STDAGs}\label{subsec:streamflow-dgp}
\begin{figure}[htp]
    \centering
    \begin{subfigure}[b]{0.75\linewidth}
        \centering
        \includegraphics[width=\linewidth]{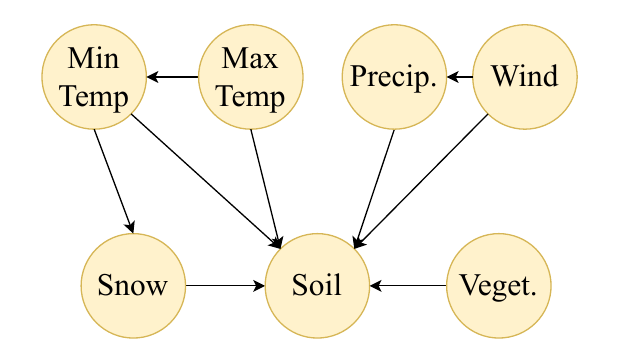}
        \caption{An example of Forcing DAG $\mathcal{G}_F=\langle(F,\mathbf{r}),\mathcal{E}_F\rangle$. A forcing DAG $\mathcal{G}_F$ captures causal relationships among forcing variables. Runoff omitted for simplicity.}
        \label{fig:forcing-dag}
    \end{subfigure}
    \vfill
    \begin{subfigure}[b]{0.75\linewidth}
        \centering
        \includegraphics[width=\linewidth]{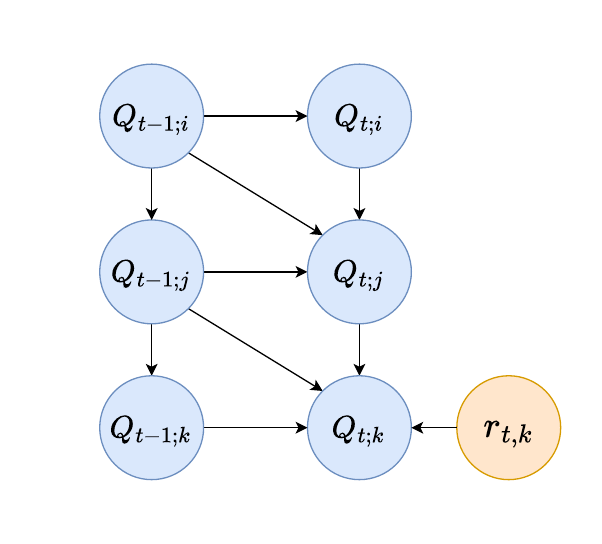}
         \captionsetup{width=\linewidth}
        \caption{Streamflow DAG $\mathcal{G}_Q=\langle(\mathbf{r}, \mathbf{Q}),\mathcal{E}_Q,\ell\rangle$ capture spatiotemporal dependencies among stations. For simplicity, we omit runoff $r$ for other stations here to illustrate the case when $Q_{t,k}$ is the target station.}
        \label{fig:streamflow-dag}
    \end{subfigure}
    \caption{CauSTream learns two causal DAGs. One instantaneous forcing DAG and an $\ell$-windowed streamflow DAG.}
    \label{fig:caustream-dag}
\end{figure}

Motivated by the VIC–CaMa-Flood framework (Fig.~\ref{fig:vic-cmf}), we design two STDAGs (Fig.~\ref{fig:caustream-dag}) in CauSTream: a forcing DAG ($\mathcal{G}_{F}$) and a routing DAG ($\mathcal{G}_Q$).

\begin{itemize}
    \item $\mathcal{G}_F=\langle(F,\mathbf{r}),\mathcal{E}_F\rangle$: an instantaneous DAG that captures causal dependencies among meteorological forcings $F$ at each timestep $t$, with runoff $\mathbf{r}$ serving as the sink node that aggregates the effects of local forcings.
    \item $\mathcal{G}_Q=\langle(\mathbf{r}, \mathbf{Q}),\mathcal{E}_Q,\ell\rangle$: an $\ell$-windowed DAG~\cite{gong2024causal}, capturing both instantaneous and lagged causal influences among streamflows $\mathbf{Q}$ and runoffs $\mathbf{r}$ across stations.
\end{itemize}

\begin{remark}
\textbf{Remark:} 
The forcing DAG $\mathcal{G}_F$ models how meteorological forcings interact to generate local runoff at each station, while the routing DAG $\mathcal{G}_Q$ governs how this runoff, together with upstream flows, propagates downstream over time.
\end{remark}

For each station $k$ at time $t$, we model the entire water basin system with a structural causal model (SCM) defined by the aforementioned graphs.

\begin{equation}\label{eq:total_dgp}
\begin{array}{r@{\quad}l}
    &F_{i, t,k} = f_{F}(\mathbf{F}_{t,k}) + \boldsymbol{\varepsilon}^{(r)}_{t,k} \quad i\in \{1, \dots, d_f\} \\
    &\mathbf{r}_{t,k} = f_{r}(\mathbf{F}_{t,k}) + \boldsymbol{\varepsilon}^{(r)}_{t,k}, \\
    &Q_{t,k} = f_{Q}\big(\mathbf{r}, \mathbf{Q_{t}}\big) + \varepsilon^{(Q)}_{t,k},
\end{array}
\end{equation}

where $f_*$ are nonlinear mixing functions, and $\boldsymbol{\varepsilon}^{(r)}$ and $\varepsilon^{(Q)}$ are mutually independent exogenous noises. We now describe each stage in detail.

\paragraph{Runoff Generation} 

Each forcing component $F_{i,t,k}$ is generated as a nonlinear function of other forcings at the same station, following the causal structure of the instantaneous forcing DAG $\mathcal{G}_F$ (Fig.~\ref{fig:forcing-dag}). The forcings $\mathbf{F}_{t,k}\in\mathbb{R}^{d_f}$ include meteorological and quasi-static variables (e.g., precipitation, temperature, land characteristics).  

At each location $k$, the local runoff $\mathbf{r}_{t,k}\in\mathbb{R}^{d_r}$ is then produced from the concurrent forcings through a nonlinear transformation $f_r(\mathbf{F}_{t,k})$, representing the hydrological process that converts atmospheric inputs into surface and subsurface flow. In this stage, forcings act as the causal parents of runoff, i.e., $pa(r_{t,k}) = \mathbf{F}_{t,k}$.

\begin{remark}
\textbf{Remark:} 

The forcing DAG $\mathcal{G}_F$ is modeled as instantaneous because meteorological forcings are typically updated daily, and most physical delays (e.g., soil moisture retention) are better captured in the runoff dynamics rather than as lagged forcing edges.  
For identifying the routing DAG $\mathcal{G}_Q$, we treat the runoff vector $\mathbf{r}_t = \{r_{t,k}\}_{k=1}^N$ as a multi-dimensional source variable that mediates between $\mathcal{G}_F$ and $\mathcal{G}_Q$. This abstraction simplifies causal discovery in $\mathcal{G}_Q$ while preserving independent noise terms $\varepsilon^{(r)}_{t,k}$ for each station’s runoff.
\end{remark}

\paragraph{Streamflow Generation} 
For each station $k$, streamflow $Q_{t,k}$ is modeled as a function of its causal parents in the routing DAG $\mathcal{G}_Q$. The parent set includes the local runoff and upstream streamflow within a lag window of length $L$: 
\[
pa(Q_{t,k}) = \{\mathbf{r}_{t,k}\} \cup \{Q_{t-\ell,j} \mid \mathbf{M}_{kj}=1,\;\ell \in \{0,\dots,L\}\},
\]
where $\mathbf{M}\in\{0,1\}^{N\times N}$ is a binary adjacency mask encoding the river network connectivity. This mask ensures that only hydrologically connected upstream stations contribute to the generation of $Q_{t,k}$, preserving both spatial sparsity and physical interpretability (Fig.~\ref{fig:streamflow-dag}).

\begin{remark}
\textbf{Remark:} Lagged edges in $\mathcal{G}_Q$ are essential for representing flow travel times that extend beyond daily resolution. For example, flow from Whitney to Richmond on the Brazos River can take 92–213 hours (approximately 4–9 days) depending on discharge~\cite{mills1970time}, with similar multi-day delays observed in the Colorado-TX and Upper Mississippi basins~\cite{waldon1998time}.  
The maximum lag $L$ should therefore reflect basin-specific travel times. In this study, we set $L=1$ as a minimal yet effective choice that captures essential routing dependencies while remaining computationally efficient.
\end{remark}

\section{\texorpdfstring{\caustream: Forecasting by Learning}{CauStream: Forecasting by Learning}}

\subsection{Theoretical Foundations and Identifiability}\label{subsec:identifiability}
We formalize identifiability for the SCM \eqref{eq:total_dgp} by linking structural equation models (SEMs) to nonlinear ICA. The key idea is that, under appropriate assumptions, the causal adjacency Jacobian of an SEM can be analytically recovered from the mixing Jacobian of its equivalent ICA representation.

\paragraph{Jacobian Notations.}
We define two key Jacobian matrices. The \textbf{causal adjacency Jacobian}, $J_g(\cdot)$, represents direct causal influences in an SEM, where its support corresponds to the causal DAG. The \textbf{mixing Jacobian}, $J_m(\cdot)$, represents the derivatives of observed variables with respect to their ultimate independent noise sources in the equivalent ICA model.

\begin{assumption}[Functional Faithfulness]\label{asm:func-faith}
For any variables $v_i, v_j \in \mathcal{V}$ in the causal graph, 
\[
v_j \in pa(v_i) \quad \iff \quad \frac{\partial f_i}{\partial v_j} \not\equiv 0.
\]
\end{assumption}

\begin{assumption}[Independent Exogenous Noise]\label{asm:indep-noise}
Let $\boldsymbol{\varepsilon} = \{\boldsymbol{\varepsilon}^{(F)}_t, \boldsymbol{\varepsilon}^{(r)}_t, \boldsymbol{\varepsilon}^{(Q)}_t\}$. Then
\[
p(\boldsymbol{\varepsilon}) = \prod_i p(\varepsilon_i),
\]
i.e., all exogenous noise terms are mutually independent across stages and stations.
\end{assumption}

\begin{assumption}[Non-Gaussian Exogenous Noise]\label{asm:non-gaussian}
For each independent noise component $\varepsilon_i \in \boldsymbol{\varepsilon}$, 
\[
\varepsilon_i \sim p_i \quad \text{with $p_i$ non-Gaussian.}
\]
\end{assumption}

\begin{remark}
\textbf{Justification of Assumptions.} 
Assumption~\ref{asm:func-faith} excludes degenerate cases where a parent has no functional effect despite being connected, ensuring that the support of the Jacobian corresponds to the true DAG. 
Assumption~\ref{asm:indep-noise} is reasonable in hydrology, since meteorological variability, runoff generation uncertainty, and routing noise arise from distinct physical processes. 
Assumption~\ref{asm:non-gaussian} is required for ICA identifiability, and is realistic given that hydrologic noise sources (e.g., precipitation and flow fluctuations) typically exhibit heavy tails and skewness rather than Gaussian behavior.
\end{remark}

\paragraph{Identifiability of the Forcing-Level DAG ($\mathcal{G}_F$)}
The forcing variables $\mathbf{F}_t$ are generated by an SEM structured by $\mathcal{G}_F$, with independent exogenous noises $\boldsymbol{\varepsilon}^{(F)}_t$. Define the source vector as $\mathbf{S}_t^{(F)} = \boldsymbol{\varepsilon}^{(F)}_t$.

\begin{corollary}[Forcing DAG Identifiability]
Let the SEM for forcing variables be $\mathbf{F}_t = f_F(\mathbf{F}_t,\mathbf{S}_t^{(F)})$ and its equivalent ICA model be $\mathbf{F}_t = f_m(\mathbf{S}_t^{(F)})$. Under Assumptions~\ref{asm:func-faith}–\ref{asm:non-gaussian}, the mixing Jacobian $J_m(\mathbf{S}_t^{(F)})$ is identifiable, and the causal adjacency Jacobian satisfies
\begin{equation}
J_g(\mathbf{F}_t) = I - D_m(\mathbf{S}_t^{(F)})\, J_m^{-1}(\mathbf{S}_t^{(F)}),
\end{equation}
where $D_m$ is the diagonal matrix containing the diagonal elements of $J_m$.
\end{corollary}

\paragraph{Identifiability of the Station-Level DAG ($\mathcal{G}_Q$)}
A similar principle applies to the streamflow generation process. The ultimate independent sources for this system are the runoff noises $\boldsymbol{\varepsilon}^{(r)}_t$ and the streamflow noises $\boldsymbol{\varepsilon}^{(Q)}_t$, which we denote collectively as $\mathbf{S}_t = \{\boldsymbol{\varepsilon}^{(r)}_t,\boldsymbol{\varepsilon}^{(Q)}_t\}$. Runoff $\mathbf{r}_t$ is a deterministic transformation of $\mathbf{F}_t$ and $\boldsymbol{\varepsilon}^{(r)}_t$, and thus inherits its independence from the exogenous sources.

\begin{corollary}[Streamflow DAG Identifiability]
Let the streamflow SEM be $\mathbf{Q}_t = f_Q(\mathbf{Q}_t,\mathbf{r}_t) + \boldsymbol{\varepsilon}^{(Q)}_t$, with $\mathbf{r}_t = f_r(\mathbf{F}_t) + \boldsymbol{\varepsilon}^{(r)}_t$. Under Assumptions~\ref{asm:indep-noise}–\ref{asm:non-gaussian}, $\mathbf{Q}_t$ admits an equivalent nonlinear ICA representation $\mathbf{Q}_t = g'_m(\mathbf{S}_t)$, where the mixing Jacobian $J'_m(\mathbf{S}_t)$ is identifiable. The causal adjacency Jacobian is given by
\begin{equation}
J_g(\mathbf{Q}_t) = I - D'_m(\mathbf{S}_t)\,(J'_m)^{-1}(\mathbf{S}_t),
\end{equation}
and its support, masked by the DEM-derived matrix $\mathbf{M}$, identifies $\mathcal{G}_Q$. Assumption~\ref{asm:func-faith} ensures that lagged edges are represented in the support.
\end{corollary}

\begin{remark}
\textbf{Proof Sketch.} For both the forcing DAG $\mathcal{G}_F$ and the routing DAG $\mathcal{G}_Q$, the corresponding SEM can be rewritten as a nonlinear ICA model that mixes a vector of independent exogenous sources $\mathbf{S}_t$ (with $\mathbf{S}_t^{(F)}=\boldsymbol{\varepsilon}^{(F)}_t$ for forcings and $\mathbf{S}_t^{(Q)}=\{\boldsymbol{\varepsilon}^{(r)}_t,\boldsymbol{\varepsilon}^{(Q)}_t\}$ for streamflow). By Assumptions~\ref{asm:indep-noise}–\ref{asm:non-gaussian}, the mixing Jacobian $J_m(\mathbf{S}_t)$ is identifiable up to scaling and permutation. Acyclicity and functional faithfulness (Assumption~\ref{asm:func-faith}) remove these ambiguities, ensuring that the support of the causal Jacobian $J_g$ coincides with the true DAG structure. Differentiating the SEM equations yields the closed-form relation between $J_g$ and $J_m$ in Corollaries 1 and 2. For the routing DAG $\mathcal{G}_Q$, an additional masking by the DEM matrix $\mathbf{M}$ enforces hydrologic feasibility, ruling out edges inconsistent with upstream–downstream connectivity.
\end{remark}

Based on these theoretical foundations, we develop CauSTream, a deep generative framework that combines causal discovery with streamflow forecasting. The architecture consists of three components: (i) \emph{Causal Representation Learning}, which disentangles the main physical processes, (ii) \emph{Causal Graph Learning}, which infers the underlying causal structures, and (iii) \emph{Spatiotemporal Forecasting}, which uses the learned structure to predict future streamflow.

\subsection{Causal Representation Learning}
Causal representation learning consists of two parts. 
First, to discover the forcing graph $\mathcal{G}_F$, we employ a variational autoencoder (VAE) to model the generative process $\mathbf{F}_t = f_F(\boldsymbol{\varepsilon}^{(F)}_t)$. 
The encoder $E_F$ learns the approximate posterior $q_\phi(\boldsymbol{\varepsilon}^{(F)}_t | \mathbf{F}_t)$, while the decoder $D_F$ learns the likelihood $p_\theta(\mathbf{F}_t | \boldsymbol{\varepsilon}^{(F)}_t)$. 
The objective maximizes the Evidence Lower Bound (ELBO):
$$
\mathcal{L}_{\text{ELBO}} = \mathbb{E}_{q_\phi}[\log p_\theta(\mathbf{F}_t|\boldsymbol{\varepsilon}^{(F)}_t)] - D_{KL}(q_\phi(\boldsymbol{\varepsilon}^{(F)}_t|\mathbf{F}_t) \,||\, p(\boldsymbol{\varepsilon}^{(F)}_t)),
$$
where the first term corresponds to the reconstruction loss (implemented as Mean Squared Error) and the second to the KL divergence, which regularizes the inferred noise distribution. 
Following identifiability results in nonlinear causal representation learning, we adopt a Laplace prior for $p(\boldsymbol{\varepsilon}^{(F)}_t)$.

Second, the runoff generation module produces runoff embeddings $\mathbf{r}_t = f_r(\mathbf{F}_t)$, which serve as causal features for streamflow forecasting. 
By default, CauSTream (\textbf{CauSTream-Shared}) uses a single shared network for $f_r$, assuming a spatially homogeneous runoff process across all stations. 
To accommodate spatial heterogeneity, we introduce a local variant (\textbf{CauSTream-Local}) where $f_r$ is parameterized by a hypernetwork~\cite{ha2017hypernetworks}, enabling station-specific parameters that adapt to local hydrological conditions—analogous to the calibration step in traditional process-based models.

\subsection{Causal Graph Learning}
Causal graph learning estimates the structures of both $\mathcal{G}_F$ and $\mathcal{G}_Q$ by applying structural penalties to the Jacobians of their respective generative functions. 
The forcing graph $\mathcal{G}_F$ is extracted from the Jacobian of the forcing decoder ($D_F$), while the streamflow graph $\mathcal{G}_Q$ is derived from the Jacobian of the forecasting component. 
Learning is guided by two regularization terms:
\begin{itemize}
    \item \textbf{Sparsity Loss ($\mathcal{L}_{\text{sparse}}$):} an L1 penalty on the masked Jacobians to encourage sparse graphs:
    $$\mathcal{L}_{\text{sparse}} = \|J_g(\mathbf{F}_t)\|_1 + \|\mathbf{M} \odot J_g(\mathbf{Q}_t)\|_1$$
    \item \textbf{Acyclicity Loss ($\mathcal{L}_{\text{DAG}}$):} a smooth penalty to ensure the learned graphs are valid DAGs \cite{zheng2018dags}:
    $$\mathcal{L}_{\text{DAG}} = \text{tr}\left(e^{\mathbf{A}_F \odot \mathbf{A}_F}\right) - d_f + \text{tr}\left(e^{\mathbf{A}_Q \odot \mathbf{A}_Q}\right) - N,$$
    where $\mathbf{A}_F$ and $\mathbf{A}_Q$ are the learned weighted adjacency matrices.
\end{itemize}

\subsection{Spatiotemporal Forecasting}
This component performs supervised forecasting with a history window of length $\ell$ as input and the next $H$ timesteps as output. As in CSF, we use a standard STGCN where the learned streamflow adjacency matrix ($\mathbf{A}_Q$) guides message passing between stations. The model predicts the next-step streamflow from historical forcings and streamflow, then rolls out iteratively for $h=1,\dots,H$ by feeding each prediction back into the context. The training loss is the horizon-averaged MAE:
\begin{equation}
\mathcal{L}_{\text{forecast}}=\frac{1}{H}\sum_{h=1}^{H}\left\lvert \mathbf{Q}_{t+h}-\hat{\mathbf{Q}}_{t+h}\right\rvert.
\end{equation}

\begin{remark}
    \textbf{Remark:} The default training setup uses past forcings and streamflow over a history window of length $\ell$ to predict the next $H$ steps. If reliable meteorological forecasts are available, we can instead condition directly on the forecasted forcings over the horizon and predict streamflow aligned with the same future period, i.e., current forcings $\to$ current horizon.
\end{remark}

\subsection{Training Objective and Optimization}
The complete model is trained end-to-end by optimizing a weighted sum of all loss components:
$$ \mathcal{L}_{\text{total}} = \mathcal{L}_{\text{forecast}} + \lambda_{\text{elbo}}\mathcal{L}_{\text{ELBO}} + \lambda_{\text{sparse}}\mathcal{L}_{\text{sparse}} + \lambda_{\text{DAG}}\mathcal{L}_{\text{DAG}}$$
We employ a curriculum learning strategy~\cite{bengio2009curriculum}, first training on the forecast and ELBO objectives before gradually annealing the weights of the structural penalties. The optimization is performed using the Adam optimizer \cite{kingma2014adam} with gradient clipping.

\subsection{Forecasting Inference}
After training, CauSTream provides two outputs: the learned causal graphs ($\mathbf{A}_F, \mathbf{A}_Q$) for analysis, and a forecasting engine for prediction. 
Given a historical window of data, the model uses the trained runoff generation component to produce runoff embeddings. 
These embeddings, together with the learned graph $\mathbf{A}_Q$, are then passed to the spatiotemporal forecasting component to generate streamflow predictions.

\section{Experiments}\label{experiments}

\subsection{Datasets}
Our experiments are conducted on datasets from three large-scale, hydrologically diverse river basins: the Brazos (73 stations), Colorado (10 stations), and Upper Mississippi (8 stations). These basins were chosen specifically for their different climatic and hydrological regimes, which allows for a robust evaluation of the model's generalizability.

All datasets span from October 1, 1973, to September 27, 1977. The daily meteorological forcings, including precipitation, minimum/maximum temperature, and wind speed, are sourced from the Livneh et al. dataset \cite{livneh2013long} for the conterminous United States. The corresponding daily streamflow observations for all stations are obtained from the U.S. Geological Survey's (USGS) National Water Information System (NWIS).

Additionally, to validate the model's internal representations and learned graphs, the Brazos basin dataset has two additional datasets. A ground-truth \emph{river flow graph}, derived from elevations (DEMs), is used to evaluate the learned streamflow graph, $\mathcal{G}_Q$. Furthermore, we use simulated runoff values from a calibrated VIC model, driven by the identical forcings, to serve as a physical proxy for evaluating our learned runoff embeddings, $\mathbf{r}_t$.

\subsection{Experiment Settings}\label{subsec:exp_settings}

To evaluate the CauSTream framework, our experimental design is structured to provide comprehensive insights into key areas of analysis that will be detailed in the Results section: (1) forecasting performance against state-of-the-art baselines, (2) the physical plausibility of the learned causal graphs, and (3) the quality of the internal runoff representations.

\paragraph{Baselines and Implementation Details.}
We benchmark CauSTream against a comprehensive suite of models, including ConvLSTM~\cite{shi2015convolutional}, STGCN~\cite{yu2017spatio}, MTGNN~\cite{wu2020connecting}, TCDF~\cite{nauta2019causal}, and CSF~\cite{wan2024csf}. To ensure a fair comparison, no prior knowledge is provided to any of the models\footnote{The original CSF uses the river network as the prior knowledge.}. For CauSTream, we set the runoff embedding dimension to $d_r=2$. Hyperparameters for all baselines are tuned to maximize the Nash-Sutcliffe Efficiency (NSE) on a validation set. All experiments are conducted on machines with NVIDIA A100 GPUs using PyTorch version 2.6.

\paragraph{Forecasting Tasks.}
We evaluate models under short-, medium-, and long-range settings. 
The short-, medium-, and long-range cases use 7/1, 14/3, and 28/7 day input/output windows, respectively.
These choices are operationally oriented~\cite{WMO1364} and consistent with typical streamflow travel times~\cite{mills1970time,waldon1998time}, making them both practical and hydrologically justified.
All models are trained with a rolling pipeline, where 1-day forecasts are autoregressively extended to multi-step predictions.

\begin{table*}[ht!]
\centering
\caption{Forecasting Performance in the Brazos Basin (1973 - 1977). The best performance in each column is in \textbf{bold}, and the second-best is \underline{underlined}.}
\label{tab:forecasting_results}
\begin{tabular}{@{}lccccccccccccccc@{}}
\toprule
\textbf{Model} & \multicolumn{4}{c}{\textbf{Short Range}} & \phantom{} & \multicolumn{4}{c}{\textbf{Medium Range}} & \phantom{} & \multicolumn{4}{c}{\textbf{Long Range}} \\
\cmidrule{2-5} \cmidrule{7-10} \cmidrule{12-15}
 & \textbf{NSE $\uparrow$} & \textbf{KGE $\uparrow$} & \textbf{VE $\uparrow$} & \textbf{$\rho$ $\uparrow$} & & \textbf{NSE $\uparrow$} & \textbf{KGE $\uparrow$} & \textbf{VE $\uparrow$} & \textbf{$\rho$ $\uparrow$} & & \textbf{NSE $\uparrow$} & \textbf{KGE $\uparrow$} & \textbf{VE $\uparrow$} & \textbf{$\rho$ $\uparrow$} \\
\midrule
Conv-LSTM & 0.59 & 0.62 & 0.62 & 0.64 & & 0.66 & 0.62 & 0.61 & 0.64 & & 0.53 & 0.51 & 0.57 & 0.52 \\
TCDF & 0.74 & 0.71 & 0.73 & 0.74 & & 0.64 & 0.63 & 0.67 & 0.63 & & 0.61 & 0.57 & 0.62 & 0.61 \\
MTGNN & 0.73 & 0.69 & 0.71 & 0.69 & & 0.63 & 0.62 & 0.63 & 0.64 & & 0.60 & 0.63 & 0.61 & 0.62 \\
STGCN & 0.77 & 0.79 & 0.75 & 0.81 & & 0.74 & 0.72 & 0.75 & 0.71 & & 0.66 & 0.62 & 0.64 & 0.65 \\
CSF & {0.82} & 0.83 & 0.82 & 0.83 & & 0.76 & 0.80 & 0.79 & 0.81 & & 0.71 & \underline{0.69} & 0.64 & 0.73 \\
\midrule
CauSTream (Shared) & \underline{0.87} & \underline{0.84} & \textbf{0.87} & \underline{0.89} & & \underline{0.79} & \textbf{0.85} & \underline{0.83} & \textbf{0.88} & & \underline{0.72} & 0.67 & \underline{0.71} & \underline{0.74} \\
CauSTream (Local) & \textbf{0.89} & \textbf{0.86} & \underline{0.86} & \textbf{0.92} & & \textbf{0.84} & \underline{0.84} & \textbf{0.85} & \underline{0.87} & & \textbf{0.77} & \textbf{0.72} & \textbf{0.74} & \textbf{0.78} \\
\bottomrule
\end{tabular}
\end{table*}

\subsection{Evaluation}
We design a comprehensive set of experiments to evaluate the CauSTream framework. The evaluation is structured to validate the model's forecasting accuracy, the correctness of the discovered causal graphs, and the quality of the learned internal representations.

\subsubsection{Forecasting Performance}
The primary evaluation of the model's predictive capability is compared against a suite of baselines across multiple basins and forecast horizons. Performance is measured using several metrics common in hydrologic assessment \cite{WMO1364}, providing comprehensive insights into the accuracy and reliability of the model's predictions.

\paragraph{Nash-Sutcliffe Efficiency (NSE) \cite{nash1970river}} The NSE measures how well the predicted values match the observed data relative to the mean. It is defined as:
\[
\text{NSE} = 1 - \frac{\sum_{t=1}^{T} (y_t - \hat{y}_t)^2}{\sum_{t=1}^{T} (y_t - \bar{y})^2},
\]
where $ y_t $ is the observed value, $ \hat{y}_t $ is the predicted value, and $ \bar{y} $ is the mean of the observed values. An NSE value of 1 indicates a perfect match.

\paragraph{Kling-Gupta Efficiency (KGE) \cite{gupta2011typical}} The KGE is designed to provide a more balanced assessment of model performance by considering the correlation, bias, and variability ratio between observed and predicted values. It is defined as:
\[
\text{KGE} = 1 - \sqrt{(r-1)^2 + (\beta-1)^2 + (\gamma-1)^2}
\]
where $ r $ is the linear correlation coefficient, $ \beta $ is the bias ratio, and $ \gamma $ is the variability ratio. A value of 1 indicates optimal performance.

\paragraph{Volumetric Efficiency (VE) \cite{criss2008nash}} The VE assesses the accuracy of model predictions by calculating the absolute volume difference between observed and predicted values. VE is defined as:
\[
VE = 1 - \frac{\sum |\hat{y}_t - y_t|}{\sum y_t}
\]
A VE value of 1 indicates perfect volumetric agreement.

\paragraph{Pearson Correlation Coefficient ($\rho$)} The Pearson correlation coefficient $ \rho $ between the observed values $ y $ and predicted values $ \hat{y} $ is defined as
$$
\rho = \frac{\text{cov}(y, \hat{y})}{\sigma(y) \sigma({\hat{y}})}.
$$

\subsubsection{Runoff Embedding Alignment}
To evaluate the physical plausibility of the learned runoff embedding $\mathbf{r}_t$, we compare it against runoff values, $\tilde{\mathbf{r}}_t$, simulated by the established VIC hydrological model. These simulations are driven by the identical meteorological forcings used as input to our model.

In addition to a direct comparison is insufficient, we also quantify the alignment using the following two robust metrics.

\paragraph{Coefficient of Determination ($R^2$)}
We first account for the potential nonlinear relationship by fitting a kernel ridge regressor, $\hat{g}_k$, which learns the mapping from our embedding to the VIC runoff ($r_{t,k} \mapsto \tilde{r}_{t,k}$) for each station $k$. We then compute the $R^2$ of this regression on a test set to measure the explained variance:
$$
R^{2}_{k} = 1 - \frac{\sum_{t=1}^{T}\bigl(\tilde{r}_{t,k} - \hat{g}_k(r_{t,k})\bigr)^{2}}{\sum_{t=1}^{T}\bigl(\tilde{r}_{t,k} - \bar{\tilde{r}}_{k}\bigr)^{2}}
$$
The final score is the average $R^2$ across all stations.

\paragraph{Mean Rank Correlation (MCC)}
This metric uses the Spearman's rank correlation coefficient ($\rho_s$) to measure the component-wise monotonic alignment between the two representations. We compute the MCC by matching each dimension of our learned embedding to the VIC runoff series with which it has the highest absolute rank correlation, and averaging these best-match correlations:
\[
\mathrm{MCC} = \frac{1}{d_r}\sum_{i=1}^{d_r} \max_{j}\bigl|\rho_{s}(r_{\cdot,i}, \tilde{r}_{\cdot,j})\bigr|
\]
where $d_r$ is the dimension of the runoff embedding. A score close to 1 indicates a strong alignment.

\begin{table*}[ht!]
\centering
\caption{Forecasting Performance (NSE) across Three Basins.}
\label{tab:basin_results}
\resizebox{\textwidth}{!}{%
\begin{tabular}{@{}l ccc c ccc c ccc@{}}
\toprule
\textbf{Model} & \multicolumn{3}{c}{\textbf{Short Range}} & \phantom{} & \multicolumn{3}{c}{\textbf{Medium Range}} & \phantom{} & \multicolumn{3}{c}{\textbf{Long Range}} \\
\cmidrule(lr){2-4} \cmidrule(lr){6-8} \cmidrule(lr){10-12}
 & \textbf{Brazos} & \textbf{Colorado} & \textbf{Mississippi} & & \textbf{Brazos} & \textbf{Colorado} & \textbf{Mississippi} & & \textbf{Brazos} & \textbf{Colorado} & \textbf{Mississippi} \\
\midrule
Conv-LSTM & 0.59 & 0.57 & 0.65 & & 0.66 & 0.60 & 0.62 & & 0.53 & 0.49 & 0.57 \\
TCDF & 0.74 & 0.65 & 0.66 & & 0.64 & 0.59 & 0.62 & & 0.61 & 0.53 & 0.59 \\
MTGNN & 0.73 & 0.61 & 0.73 & & 0.63 & 0.62 & 0.62 & & 0.60 & 0.61 & 0.62 \\
STGCN & 0.77 & 0.68 & 0.76 & & 0.74 & 0.71 & 0.70 & & 0.66 & 0.63 & 0.64 \\
CSF & 0.82 & 0.72 & 0.77 & & 0.76 & 0.69 & 0.74 & & 0.71 & 0.64 & 0.71 \\
\midrule
CauSTream (Shared) & \underline{0.87} & \underline{0.76} & \underline{0.81} & & \underline{0.79} & \underline{0.74} & \underline{0.77} & & \underline{0.72} & \underline{0.73} & \textbf{0.75} \\
CauSTream (Local) & \textbf{0.89} & \textbf{0.80} & \textbf{0.86} & & \textbf{0.81} & \textbf{0.77} & \textbf{0.82} & & \textbf{0.77} & \textbf{0.74} & \underline{0.73} \\
\bottomrule
\end{tabular}
}
\end{table*}

\subsection{Results}
\begin{figure}[htp]
    \centering
    \begin{subfigure}[b]{.7\linewidth}
        \centering
        \includegraphics[width=\linewidth]{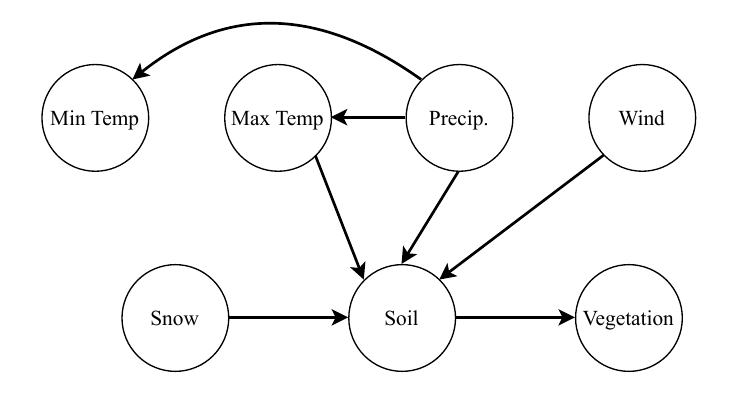}
        \caption{Brazos Basin}
        \label{fig:brazos_forcing}
    \end{subfigure}
    \vfill
    \begin{subfigure}[b]{.7\linewidth}
        \centering
        \includegraphics[width=\linewidth]{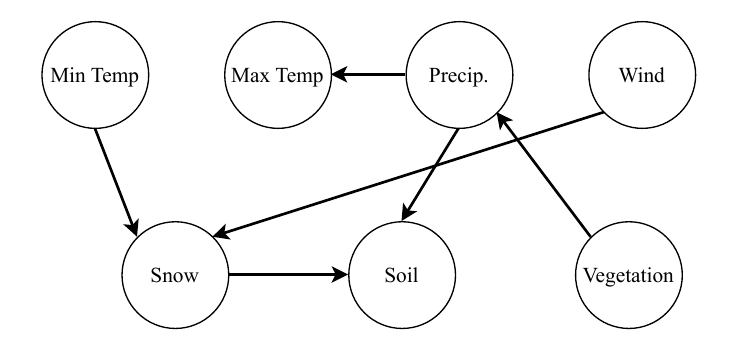}
        \caption{Mississippi Basin}
        \label{fig:mississippi_forcing}
    \end{subfigure}
    \caption{Aggregated forcing graphs ($\mathcal{G}_F$) for two stations located in the Brazos and Upper Mississippi basins.}
    \label{fig:forcing_graphs}
\end{figure}
\subsubsection{Forecasting Accuracy}

Table~\ref{tab:forecasting_results} summarizes the forecasting performance of all models across four key metrics and three forecasting ranges, while Table~\ref{tab:basin_results} provides a detailed breakdown of the primary metric (NSE) across the three distinct river basins. A clear trend emerges from these results: the CauSTream models consistently outperform all baselines.

The analysis reveals three key findings. First, the CauSTream-Local variant achieves the best performance across almost all metrics and scenarios, demonstrating the significant benefit of modeling local dynamics via the hypernetwork. Second, the framework shows strong robustness; the superior performance of the CauSTream variants holds across the Brazos, Colorado, and Upper Mississippi basins, which represent very different hydrological environments. Finally, the performance gap between CauSTream and the baseline models is most pronounced in the long-range forecasting tasks. This suggests that by learning the underlying causal structure of the system, our model is better equipped to generalize and make reliable predictions over longer time horizons where simple pattern matching is more likely to fail.

\subsubsection{Case Study: Causal Graph Discovery}

\paragraph{Forcing DAG}
As discussed in Section~\ref{subsec:streamflow-dgp}, we model forcing-runoff dynamics as instantaneous DAGs, emphasizing intra-day causal dependencies. Since DAGs are generated across locations and time, directly aggregating them would otherwise introduce bidirectional edges or cycles. To address this, we perform aggregation only over time, not across locations, because local forcings evolve differently under varying climatic conditions. We aggregate the adjacency matrices by taking their element-wise average across time steps, and then binarize the result using a consistency threshold~\eqref{eq:agg-dags}. This approach keeps the most stable and recurrent causal relationships while suppressing transient or noisy connections. The divergent dynamics observed across locations further explain why the CauSTream-Local variant performs best in experiments (Table~\ref{tab:forecasting_results}).

\begin{equation}\label{eq:agg-dags}
\bar{A}_{ij} = \frac{1}{T} \sum_{t=1}^{T} A_{ij}^{(t)}, \quad 
\tilde{A}_{ij} = 
\begin{cases}
1, & \text{if } \bar{A}_{ij} \geq \tau, \\
0, & \text{otherwise.}
\end{cases}
\end{equation}

To illustrate, Fig.~\ref{fig:forcing_graphs} compares forcing graphs from two locations: one in the Brazos Basin and another in the Upper Mississippi Basin. The Brazos site, near Lubbock in the semi-arid west, shows strong coupling between precipitation and temperature, where rainfall events induce substantial temperature drops. In contrast, the Mississippi Basin graph highlights the critical role of vegetation: dense vegetation intercepts precipitation and regulates surface runoff, while strong wind affects snow distribution. Together, these forcing graphs provide valuable insight into how local climatic forcings shape distinct runoff generation mechanisms.

\begin{figure}[ht!]
    \centering
    \begin{subfigure}[b]{0.7\linewidth}
        \centering
        \includegraphics[width=\linewidth]{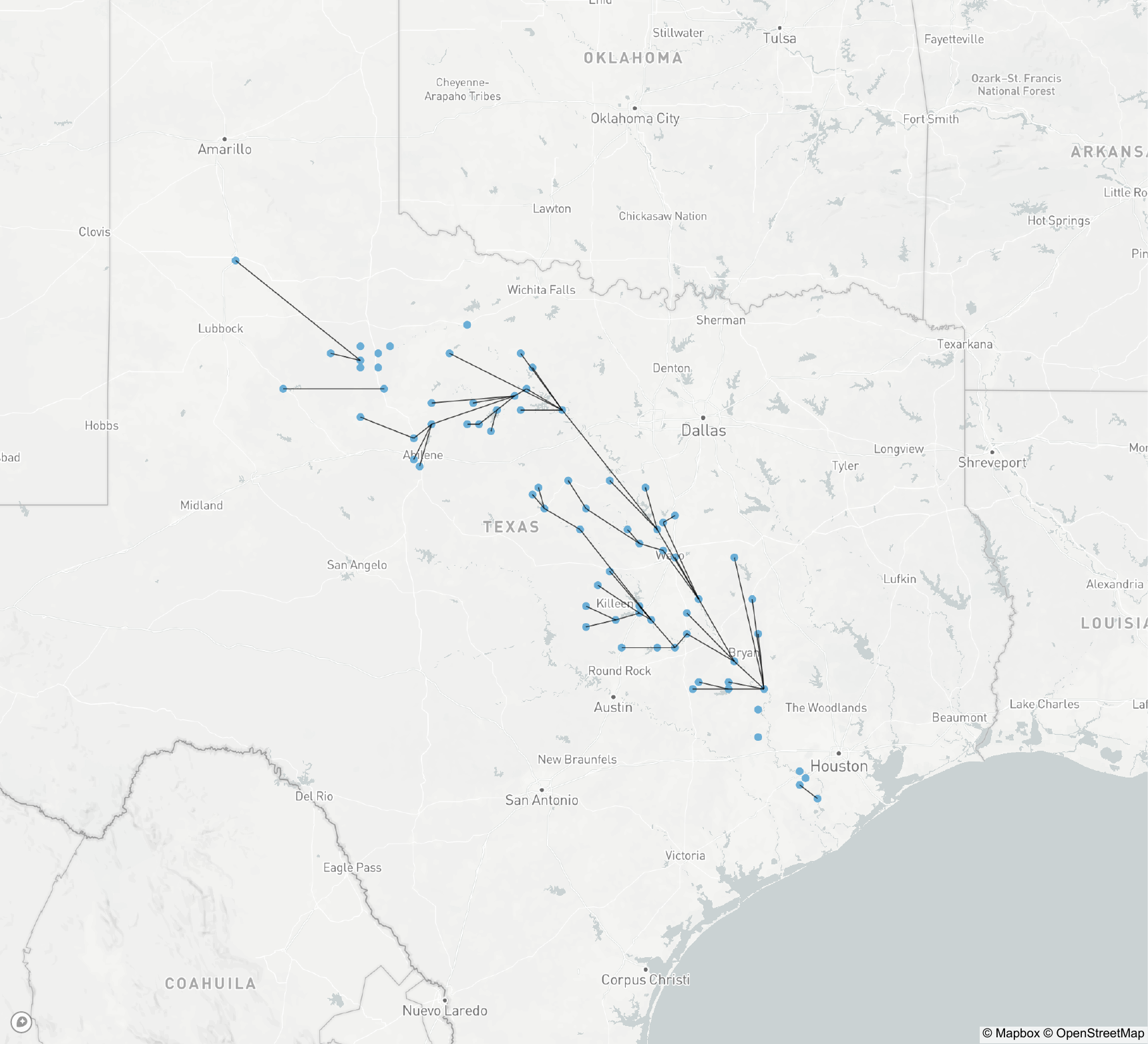}
        \caption{Ground-Truth River Network.}
        \label{fig:brazos_river_graph}
    \end{subfigure}
    \hspace{20pt}
    \begin{subfigure}[b]{0.7\linewidth}
        \centering
        \includegraphics[width=\linewidth]{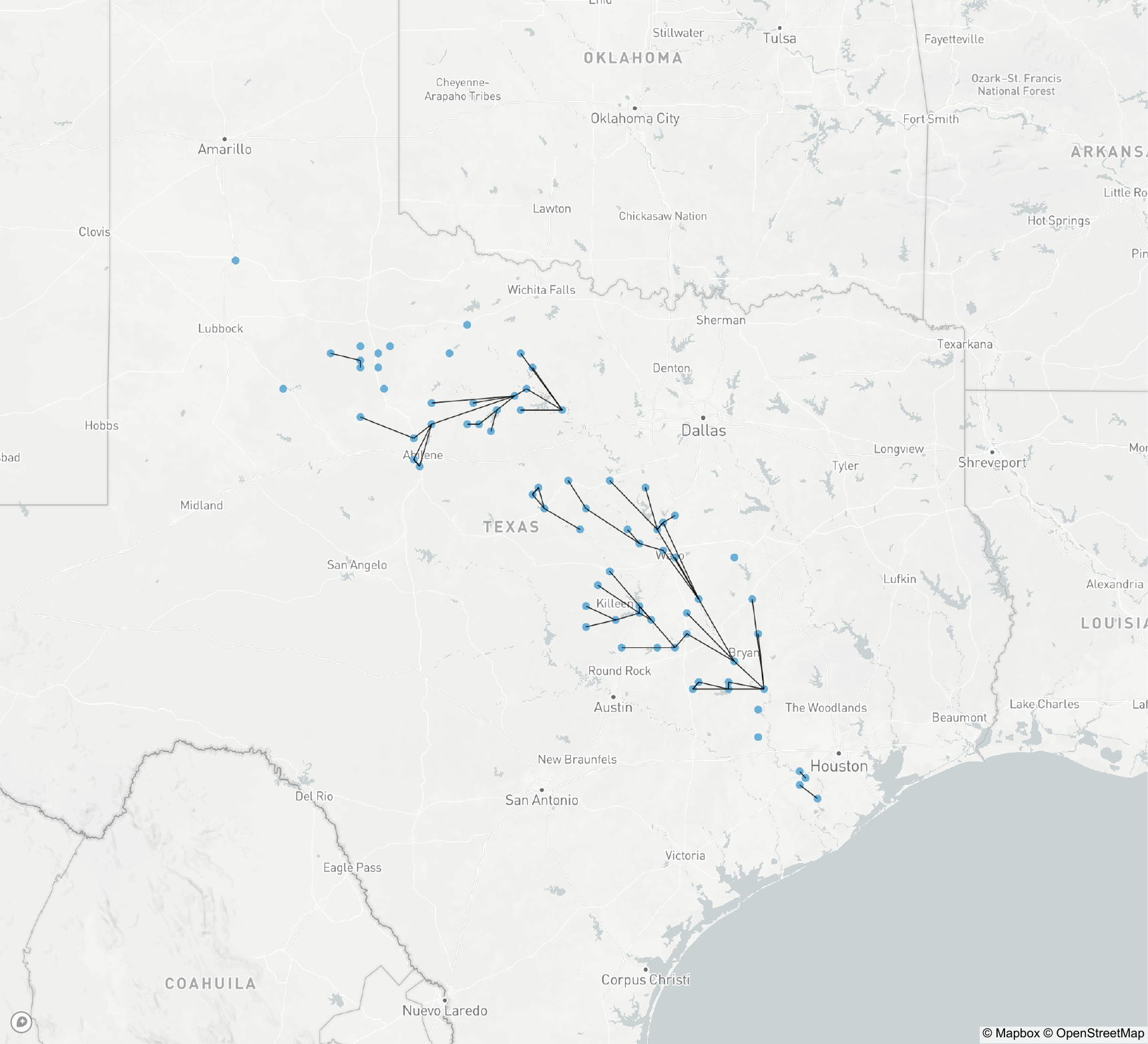}
        \caption{Learned Routing Graph ($\mathcal{G}_Q$).}
        \label{fig:brazos_streamflow_graph}
    \end{subfigure}
    \caption{Comparison between the ground-truth river network and the learned routing DAG $\mathcal{G}_Q$ for the Brazos Basin. The learned graph captures the main flow directions and exhibits clear clustering patterns that align with subbasin delineations.}
    \label{fig:streamflow_graph_comparison}
\end{figure}

\begin{table*}[htp!]
\centering
\caption{Ablation Study: Forecasting performance of CauSTream variants compared to CSF~\cite{wan2024csf}.}
\label{tab:ablation_study}
\resizebox{\textwidth}{!}{%
\begin{tabular}{@{}lccccccccccccccc@{}}
\toprule
\textbf{Model} & \multicolumn{4}{c}{\textbf{Short Range}} & \phantom{} & \multicolumn{4}{c}{\textbf{Medium Range}} & \phantom{} & \multicolumn{4}{c}{\textbf{Long Range}} \\
\cmidrule(lr){2-5} \cmidrule(lr){7-10} \cmidrule(lr){12-15}
 & \textbf{NSE $\uparrow$} & \textbf{KGE $\uparrow$} & \textbf{VE $\uparrow$} & \textbf{$\rho$ $\uparrow$} & & \textbf{NSE $\uparrow$} & \textbf{KGE $\uparrow$} & \textbf{VE $\uparrow$} & \textbf{$\rho$ $\uparrow$} & & \textbf{NSE $\uparrow$} & \textbf{KGE $\uparrow$} & \textbf{VE $\uparrow$} & \textbf{$\rho$ $\uparrow$} \\
\midrule
CauSTream (Local) - CL & 0.79 & 0.81 & 0.83 & 0.77 & & 0.75 & 0.79 & 0.76 & 0.79 & & 0.68 & 0.65 & 0.67 & 0.69 \\
CSF & 0.82 & 0.83 & 0.82 & 0.83 & & 0.76 & 0.80 & 0.79 & 0.81 & & 0.71 & \underline{0.69} & 0.64 & 0.73 \\
CauSTream (Shared) & 0.87 & 0.84 & \textbf{0.87} & \underline{0.89} & & 0.79 & \textbf{0.85} & 0.83 & \textbf{0.88} & & 0.72 & 0.67 & 0.71 & 0.74 \\
CauSTream (Local) - Forcing & \underline{0.88} & \underline{0.85} & 0.85 & 0.87 & & \underline{0.81} & \underline{0.84} & \underline{0.84} & 0.83 & & \underline{0.75} & \underline{0.69} & \underline{0.72} & \underline{0.77} \\
CauSTream (Local) & \textbf{0.89} & \textbf{0.86} & 0.86 & \textbf{0.92} & & \textbf{0.84} & \underline{0.84} & \textbf{0.85} & \underline{0.87} & & \textbf{0.77} & \textbf{0.72} & \textbf{0.74} & \textbf{0.78} \\
\bottomrule
\end{tabular}%
}
\end{table*}

\paragraph{Streamflow DAG}
A comparison between the learned streamflow graph (Fig.~\ref{fig:brazos_streamflow_graph}) and the ground-truth river network (Fig.~\ref{fig:brazos_river_graph}) for the Brazos Basin shows strong alignment. Most learned causal edges follow the downstream flow direction, suggesting that the model successfully captures the main flow patterns of the river system. The learned graph also forms clear station clusters that follow subbasin delineations~\cite{Halff2019_AppD}. The main difference is that the model tends to miss some long-distance connections between far-apart upstream and downstream stations found in the ground-truth network. This occurs because the model uses a maximum lag of one day, which limits its ability to detect delayed effects that take longer to propagate through the river network.

\subsubsection{Runoff Embedding Similarities}
Table~\ref{tab:runoff_similarity} shows the results of the runoff embedding alignment analysis. The CauSTream framework, particularly the CauSTream-Local variant, produces embeddings that are far more aligned with the VIC-simulated runoff than the baseline CSF model, achieving an MCC of 0.92 and an $R^2$ of 0.96. The superior performance of CauSTream-Local over CauSTream-Shared validates the contribution of the hypernetwork in capturing local heterogeneity. The significant gap between both CauSTream variants and CSF suggests that our approach of explicitly generating runoff embeddings for the forecasting task results in a more physically plausible representation than CSF's VAE-based method, which is optimized for input reconstruction.

\begin{table}[htp]
\centering
\caption{Runoff Embedding Similarity. Comparison of model performance on aligning learned runoff embeddings with VIC-simulated runoff.}
\label{tab:runoff_similarity}
\begin{tabular}{@{}lcc@{}}
\toprule
\textbf{Model} & \textbf{MCC} & \textbf{$R^2$} \\
\midrule
CSF & 0.43 & 0.54 \\
CauSTream (Shared) & \underline{0.77} & \underline{0.67} \\
CauSTream (Local) & \textbf{0.92} & \textbf{0.96} \\
\bottomrule
\end{tabular}
\end{table}

\subsubsection{Ablation Study}
To validate the contribution of each key component within the CauSTream architecture, we conduct an ablation study. We compare the forecasting performance of the full CauSTream-Local model against several ablated versions:
\begin{itemize}
    \item \textbf{No Causal Learning:} The model trained without the $\mathcal{L}_{\text{sparse}}$ and $\mathcal{L}_{\text{DAG}}$ losses.
    \item \textbf{No Forcing Graph Learning:} The model without the VAE-based representation learning for $\mathcal{G}_F$.
    \item \textbf{Shared vs. Local:} A direct comparison of CauSTream-Shared against CauSTream-Local to quantify the benefit of the hypernetwork.
\end{itemize}
The results (Table~\ref{tab:ablation_study}) confirm that each component of the framework provides a significant contribution. The full CauSTream-Local model consistently outperforms all ablated versions. Removing the causal learning losses results in a substantial drop in accuracy, underscoring the importance of enforcing sparse DAG structures. Similarly, the performance gap between the shared and local variants validates the effectiveness of the hypernetwork in capturing local heterogeneity. Finally, removing the forcing graph learning component also degrades performance, highlighting the benefit of creating a causally-informed runoff embedding.

\section{Conclusion, Limitations, and Future Work}\label{conclusion}

We introduced \caustream, a hydrology-guided deep generative framework that integrates causal representation learning with spatiotemporal forecasting. To the best of our knowledge, CauSTream is the first model that jointly learns causal graph structures from data and performs multi-step streamflow prediction within a single, unified framework. Extensive experiments across three distinct river basins demonstrate that CauSTream consistently outperforms a range of strong baselines. The results show that the model generalizes well across diverse hydrological regimes and that the advantage of learning causal structure becomes especially evident in long-range forecasts. Moreover, our analysis reveals that CauSTream discovers interpretable causal graphs consistent with established physical processes and learns runoff representations that exhibit strong correspondence with physically based simulations from the VIC model.

While CauSTream demonstrates strong performance and interpretability, it also opens several promising directions for future work. The current framework does not account for latent variables. This assumption is strong and may limit the model’s ability to capture unobserved confounding effects. A key direction for future research is to relax this assumption and develop a theoretically identifiable extension of CauSTream that remains valid under latent confounding.

Second, an immediate direction is to enhance the model’s scalability. Although the current implementation focuses on basin-level applications, scaling CauSTream to continental domains (e.g., the contiguous United States) would enable large-scale causal mapping of hydrological connectivity while testing the model’s efficiency and generalization under diverse climate and terrain conditions.

Finally, the learned causal graphs open new possibilities beyond forecasting. The streamflow graph provides a natural foundation for counterfactual reasoning—supporting ``what-if'' analyses that could turn CauSTream from a predictive model into an interactive decision-support system for sustainable water resource management.

\section*{Acknowledgments}
This research has been funded by NSF$\#$2311716 ``CausalBench: A Cyberinfrastructure for Causal-Learning Benchmarking for Efficacy, Reproducibility, and Scientific Collaboration" and US Army Corps of Engineers Engineering With Nature Initiative through Cooperative Ecosystem Studies Unit Agreement $\#$W912HZ-21-2-0040.

\bibliographystyle{IEEEtran}
\bibliography{references}

@article{yamazaki2011physically,
  title={A physically based description of floodplain inundation dynamics in a global river routing model},
  author={Yamazaki, Dai and Kanae, Shinjiro and Kim, Hyungjun and Oki, Taikan},
  journal={Water Resources Research},
  volume={47},
  number={4},
  year={2011},
  publisher={Wiley Online Library}
}

@article{nearing2021role,
  title={What role does hydrological science play in the age of machine learning?},
  author={Nearing, Grey S and Kratzert, Frederik and Sampson, Alden Keefe and Pelissier, Craig S and Klotz, Daniel and Frame, Jonathan M and Prieto, Cristina and Gupta, Hoshin V},
  journal={Water Resources Research},
  volume={57},
  number={3},
  pages={e2020WR028091},
  year={2021},
  publisher={Wiley Online Library}
}

@article{shi2015convolutional,
  title={Convolutional LSTM network: A machine learning approach for precipitation nowcasting},
  author={Shi, Xingjian and Chen, Zhourong and Wang, Hao and Yeung, Dit-Yan and Wong, Wai-Kin and Woo, Wang-chun},
  journal={Advances in neural information processing systems},
  volume={28},
  year={2015}
}

@article{kratzert2018rainfall,
  title={Rainfall--runoff modelling using long short-term memory (LSTM) networks},
  author={Kratzert, Frederik and Klotz, Daniel and Brenner, Claire and Schulz, Karsten and Herrnegger, Mathew},
  journal={Hydrology and Earth System Sciences},
  volume={22},
  number={11},
  pages={6005--6022},
  year={2018},
  publisher={Copernicus Publications G{\"o}ttingen, Germany}
}

@misc{chang2023artificial,
  title={Artificial intelligence techniques in hydrology and water resources management},
  author={Chang, Fi-John and Chang, Li-Chiu and Chen, Jui-Fa},
  journal={Water},
  volume={15},
  number={10},
  pages={1846},
  year={2023},
  publisher={MDPI}
}

@article{shen2018transdisciplinary,
  title={A transdisciplinary review of deep learning research and its relevance for water resources scientists},
  author={Shen, Chaopeng},
  journal={Water Resources Research},
  volume={54},
  number={11},
  pages={8558--8593},
  year={2018},
  publisher={Wiley Online Library}
}

@article{nash1970river,
  title={River flow forecasting through conceptual models part I—A discussion of principles},
  author={Nash, J Eamonn and Sutcliffe, Jonh V},
  journal={Journal of hydrology},
  volume={10},
  number={3},
  pages={282--290},
  year={1970},
  publisher={Elsevier}
}

@article{gupta2011typical,
  title={On typical range, sensitivity, and normalization of Mean Squared Error and Nash-Sutcliffe Efficiency type metrics},
  author={Gupta, Hoshin Vijai and Kling, Harald},
  journal={Water Resources Research},
  volume={47},
  number={10},
  year={2011},
  publisher={Wiley Online Library}
}

@inproceedings{yu2017spatio,
author = {Yu, Bing and Yin, Haoteng and Zhu, Zhanxing},
title = {Spatio-temporal graph convolutional networks: a deep learning framework for traffic forecasting},
year = {2018},
isbn = {9780999241127},
publisher = {AAAI Press},
booktitle = {Proceedings of the 27th International Joint Conference on Artificial Intelligence},
pages = {3634–3640},
numpages = {7},
location = {Stockholm, Sweden},
series = {IJCAI'18}
}

@article{liang1994simple,
  title={A simple hydrologically based model of land surface water and energy fluxes for general circulation models},
  author={Liang, Xu and Lettenmaier, Dennis P and Wood, Eric F and Burges, Stephen J},
  journal={Journal of Geophysical Research: Atmospheres},
  volume={99},
  number={D7},
  pages={14415--14428},
  year={1994},
  publisher={Wiley Online Library}
}

@article{criss2008nash,
  title={Do Nash values have value? Discussion and alternate proposals},
  author={Criss, Robert E and Winston, William E},
  journal={Hydrological Processes: An International Journal},
  volume={22},
  number={14},
  pages={2723--2725},
  year={2008},
  publisher={Citeseer}
}

@article{wang2021modeling,
  title={Modeling daily floods in the lancang-mekong river basin using an improved hydrological-hydrodynamic model},
  author={Wang, Jie and Yun, Xiaobo and Pokhrel, Yadu and Yamazaki, Dai and Zhao, Qiudong and Chen, Aifang and Tang, Qiuhong},
  journal={Water Resources Research},
  volume={57},
  number={8},
  pages={e2021WR029734},
  year={2021},
  publisher={Wiley Online Library}
}

@inproceedings{liu2023applying,
  title={Applying graph neural networks to improve the data resolution of stream water quality monitoring networks},
  author={Liu, Ting and Deng, Qi and Ding, Kaize and Sabo, John L and Liu, Huan and Candan, Kasim Selcuk and Muenich, Rebecca Logsdon},
  booktitle={AGU Fall Meeting Abstracts},
  volume={2023},
  pages={H42E--03},
  year={2023}
}

@inproceedings{deng2022deep,
  title={A Deep Learning Approach to Distributed Rainfall-Runoff Modeling with Spatiotemporal Interpretation.},
  author={Deng, Qi and Liu, Ting and Wei, Yuhang and Sabo, John L},
  booktitle={AGU Fall Meeting Abstracts},
  volume={2022},
  pages={IN12B--0271},
  year={2022}
}

@inproceedings{sheth2023streams,
  title={STREAMS: Towards Spatio-Temporal Causal Discovery with Reinforcement Learning for Streamflow Rate Prediction},
  author={Sheth, Paras and Mosallanezhad, Ahmadreza and Ding, Kaize and Shah, Reepal and Sabo, John and Liu, Huan and Candan, K Sel{\c{c}}uk},
  booktitle={Proceedings of the 32nd ACM International Conference on Information and Knowledge Management},
  pages={4815--4821},
  year={2023}
}

@article{christiansen2022toward,
  title={Toward causal inference for spatio-temporal data: Conflict and forest loss in Colombia},
  author={Christiansen, Rune and Baumann, Matthias and Kuemmerle, Tobias and Mahecha, Miguel D and Peters, Jonas},
  journal={Journal of the American Statistical Association},
  volume={117},
  number={538},
  pages={591--601},
  year={2022},
  publisher={Taylor \& Francis}
}

@article{nauta2019causal,
  title={Causal discovery with attention-based convolutional neural networks},
  author={Nauta, Meike and Bucur, Doina and Seifert, Christin},
  journal={Machine Learning and Knowledge Extraction},
  volume={1},
  number={1},
  pages={19},
  year={2019},
  publisher={MDPI}
}

@article{livneh2013long,
  title={A long-term hydrologically based dataset of land surface fluxes and states for the conterminous United States: Update and extensions},
  author={Livneh, Ben and Rosenberg, Eric A and Lin, Chiyu and Nijssen, Bart and Mishra, Vimal and Andreadis, Kostas M and Maurer, Edwin P and Lettenmaier, Dennis P},
  journal={Journal of Climate},
  volume={26},
  number={23},
  pages={9384--9392},
  year={2013},
  publisher={American Meteorological Society}
}

@inproceedings{wu2020connecting,
  title={Connecting the dots: Multivariate time series forecasting with graph neural networks},
  author={Wu, Zonghan and Pan, Shirui and Long, Guodong and Jiang, Jing and Chang, Xiaojun and Zhang, Chengqi},
  booktitle={Proceedings of the 26th ACM SIGKDD international conference on knowledge discovery \& data mining},
  pages={753--763},
  year={2020}
}

@article{zheng2018dags,
  title={Dags with no tears: Continuous optimization for structure learning},
  author={Zheng, Xun and Aragam, Bryon and Ravikumar, Pradeep K and Xing, Eric P},
  journal={Advances in neural information processing systems},
  volume={31},
  year={2018}
}

@InProceedings{khemakhem2020variational,
  title     = {Variational Autoencoders and Nonlinear ICA: A Unifying Framework},
  author    = {Ilyes Khemakhem and Diederik P. Kingma and Ricardo P. Monti and Aapo Hyvärinen},
  booktitle = {Proceedings of the 23rd International Conference on Artificial Intelligence and Statistics},
  series    = {Proceedings of Machine Learning Research},
  volume    = {108},
  pages     = {2207--2217},
  year      = {2020},
  month     = {26--28 Aug},
  publisher = {PMLR},
  url       = {https://proceedings.mlr.press/v108/khemakhem20a.html}
}

@inproceedings{ha2017hypernetworks,
  title     = {HyperNetworks},
  author    = {David Ha and Andrew M. Dai and Quoc V. Le},
  booktitle = {International Conference on Learning Representations},
  year      = {2017},
  url       = {https://openreview.net/forum?id=rkpACe1lx},
  note      = {Poster paper}
}

@inproceedings{kingma2014adam,
  title     = {Adam: A Method for Stochastic Optimization},
  author    = {Diederik P. Kingma and Jimmy Ba},
  booktitle = {Proceedings of the 3rd International Conference on Learning Representations (ICLR)},
  year      = {2014},
  url       = {https://arxiv.org/abs/1412.6980},
  archivePrefix = {arXiv},
  eprint    = {1412.6980}
}

@inproceedings{wan2024csf,
  title={Spatio-temporal Causal Learning for Streamflow Forecasting},
  author={Wan, Shu and Shah, Reepal and Deng, Qi and Sabo, John and Liu, Huan and Candan, K Sel{\c{c}}uk},
  booktitle={2024 IEEE International Conference on Big Data (BigData)},
  pages={6161--6170},
  year={2024},
  organization={IEEE}
}

@inproceedings{
fu2025ncdl,
title={Learning General Causal Structures with Hidden Dynamic Process for Climate Analysis},
author={Minghao Fu and Biwei Huang and Zijian Li and Yujia Zheng and Ignavier Ng and Guangyi Chen and Yingyao Hu and Kun Zhang},
booktitle={NeurIPS 2025 Workshop on CauScien: Uncovering Causality in Science},
year={2025},
url={https://openreview.net/forum?id=ZjkbK8vjr2}
}

@article{shimizu2006linear,
  title={A linear non-Gaussian acyclic model for causal discovery.},
  author={Shimizu, Shohei and Hoyer, Patrik O and Hyv{\"a}rinen, Aapo and Kerminen, Antti and Jordan, Michael},
  journal={Journal of Machine Learning Research},
  volume={7},
  number={10},
  year={2006}
}

@inproceedings{hyvarinen2019nonlinear,
  title={Nonlinear ICA using auxiliary variables and generalized contrastive learning},
  author={Hyvarinen, Aapo and Sasaki, Hiroaki and Turner, Richard},
  booktitle={The 22nd international conference on artificial intelligence and statistics},
  pages={859--868},
  year={2019},
  organization={PMLR}
}

@book{mills1970time,
  title={Time of Travel of Translatory Waves on the Brazos, Leon, and Little Rivers, Texas},
  author={Mills, W.B. and Geological Survey (U.S.) and Texas Water Development Board and Brazos River Authority},
  series={Report (Texas Water Development Board)},
  year={1970},
  publisher={Texas Water Development Board},
  url={https://books.google.com/books?id=aap\_Nu0EjO8C}
}

@article{gong2024causal,
  title={Causal discovery from temporal data: An overview and new perspectives},
  author={Gong, Chang and Zhang, Chuzhe and Yao, Di and Bi, Jingping and Li, Wenbin and Xu, Yongjun},
  journal={ACM Computing Surveys},
  volume={57},
  number={4},
  pages={1--38},
  year={2024},
  publisher={ACM New York, NY}
}

@article{waldon1998time,
  title={Time-of-travel in the lower Mississippi River: model development, calibration, and application},
  author={Waldon, Michael G},
  journal={Water environment research},
  volume={70},
  number={6},
  pages={1132--1141},
  year={1998},
  publisher={Wiley Online Library}
}

@article{zheng2022identifiability,
  title={On the identifiability of nonlinear ICA: Sparsity and beyond},
  author={Zheng, Yujia and Ng, Ignavier and Zhang, Kun},
  journal={Advances in neural information processing systems},
  volume={35},
  pages={16411--16422},
  year={2022}
}

@inproceedings{bengio2009curriculum,
  title={Curriculum learning},
  author={Bengio, Yoshua and Louradour, J{\'e}r{\^o}me and Collobert, Ronan and Weston, Jason},
  booktitle={Proceedings of the 26th annual international conference on machine learning},
  pages={41--48},
  year={2009}
}

@book{WMO1364,
  title        = {Guidelines on the Verification of Hydrological Forecasts},
  author       = {{World Meteorological Organization (WMO)}},
  year         = {2025},
  edition      = {2025 edition},
  address      = {Geneva},
  publisher    = {World Meteorological Organization},
  series       = {WMO-No. 1364},
  isbn         = {978-92-63-11364-0},
  pages        = {205},
  url          = {https://library.wmo.int/idurl/4/69478},
  note         = {Technical Commission for Weather, Climate, Hydrological, Marine and Related Environmental Services and Applications (SERCOM)}
}

@techreport{Halff2019_AppD,
  title       = {Appendix D: Hydrologic Analysis},
  author      = {{Halff Associates, Inc.}},
  institution = {Brazos River Authority},
  year        = {2019},
  month       = mar,
  note        = {Appendix to the Lower Brazos Floodplain Protection Planning Study},
  url         = {https://brazos.org/project-updates/lower-brazos-floodplain-protection-planning-study}
}

\end{document}